\documentclass{article}
\usepackage{ijcai18}

\usepackage{times}
\usepackage{soul}
\usepackage{url}
\usepackage[hidelinks]{hyperref}
\usepackage[utf8]{inputenc}
\usepackage[small]{caption}
\usepackage{indentfirst}
\usepackage{graphicx}
\title{A Minesweeper Solver Using Logic Inference, CSP and Sampling}

\author{
Yimin Tang$^1$,
Tian Jiang$^2$,
Yanpeng Hu$^3$
\\ 
$^1$ Shanghaitech University\\
$^2$ Shanghaitech University\\
$^3$ Shanghaitech University\\
tangym@shanghaitech.edu.cn  \ \ 
jiangtian@shanghaitech.edu.cn  \ \ 
huyp@shanghaitech.edu.cn
}

\begin{document}

\maketitle

\begin{abstract}
Minesweeper as a puzzle video game and is proved that it is a NPC problem. We use CSP,Logic Inference and Sampling to make a minesweeper solver and we limit our each select in 5 seconds.
\end{abstract}

\section{Introduction}

Minesweeper as a puzzle video game, has been well-known worldwide since it is preassembled in the computer system in 1995. 
The concept of AI Minesweeper solver is brought up as a consequence. From 1995 to present, multiple approaches has been applied to build up a minesweeper ranging from
use of genetic algorithms, graphical models and other learning
strategies \cite{Maznikova}. There have been previous implementations to the
CSP approach, which is the current “state of the art” method
\cite{Studholme}. Nakov and Wei \cite{Nakov} derive bounds on the complexity
of playing Minesweeper optimally. The authors formulate the
Minesweeper game as a POMDP and use enumeration to convert
the game to an MDP, while also reducing the state space. They
then use value iteration methods to solve the MDP for a 4x4
board, but the method is not scalable.

\subsection{How to Play Minesweeper }

To play a minesweeper game, the player is presented with
a rectangular grid of tiles, behind which are hidden a certain
number of randomly distributed mines. The player uncovers a tile by clicking it;
if the clicked square is not a mine, it reveals an integer (its value),
which is the number of adjacent uncovered tiles (including those
that only share a corner) that contain a mine. Given new information, the player need to decide which square to uncover next until the uncovered square is a mine in an attempt(fail) or the board is cleared(success).

All paper {\em submissions} must have a maximum of six pages, plus at most one for references. The seventh page cannot contain {\bf anything} other than references.

The length rules may change for final camera-ready versions of accepted papers, and will differ between tracks. Some tracks may include only references in the last page, whereas others allow for any content in all pages. Similarly, some tracks allow you to buy a few extra pages should you want to, whereas others don't.

\subsection{Measurement Difference between Human Beings and AI}

For a human, the fastest possible time ever completed is the only metric. So it does not matter if one loses 20 games for every game we win: only the wins count.
However, for an AI, this is clearly not a reasonable metric as a robot that can click as fast as we want to. A more generalized metric would be adapted instead:
to win as many games as possible with time limit.
We define time limit as 5secs one step here to accomplish tests in appropriate time.

\section{Difficulty}

\subsection{Infinite state space}
A difficulty for a minesweeper is that its infinite state space. The total state space is $O(C_s^m2^{m-s})$, which $C_s^m$ is the numbers of mine distribution, $2^{m-s}$ shows whether a square is revealed.

\subsection{Late influence for a decision}
Similar to the Go game, the influence of a decision might show up very late, which is another difficulty in Minesweeper.

\subsection{Uncertainty}
Unlike the Go game, there is an uncertainty in the process of solving a minesweeper. In some situation, we can not apply inference directly but make a guessing according to the probability if the square is a mine.

\subsection{Time complexity}
Minesweeper is a special case of SAT(more precisely, SAT-8).  Determining
whether a given mine configuration satisfies the board
constraints was proven to be NP-complete in 2000 \cite{Kaye}

\section{Our Strategy}
\includegraphics[width = .5\textwidth]{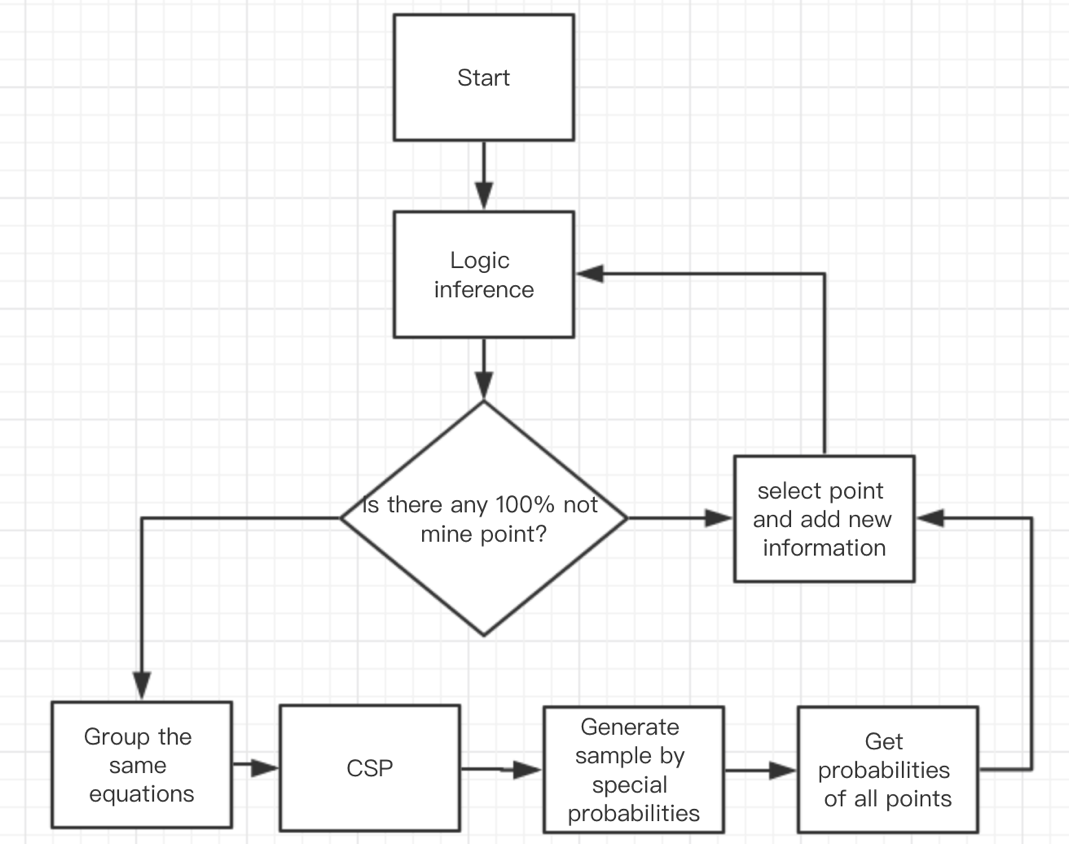}

\subsection{Logic Inference}
Each point you have chosen will be 0-9 or a mine. If it not a mine , you will get an equation like $ x_1+x_2+x_3 = 2 $. And if you have another equation like $ x_2+x_3 = 1 $ whose all variables are all contained in the first equation. You can use the first equation minus the second equation to get a shorter equation and then you can replace the longer equation with this new equation. After doing this the equation set will more laconic but equivalent. When there is a new equation, you must do the whole procedure once again until there can not be a new equation. And we think its time complexity is $O({(nm)}^2) $ where $ n,m $ is the map's length and width.

\subsection{Group the same equations}
The equation set can contain some information that has the same variables and these equations must be strongly connected. You can use Disjoint set to group these equations. we think its time complexity is $O({(nm)}) $ where $n,m$ is the map's length and width.

\subsection{CSP}
In this step, We should know how many mines are in the group can satisfy all conditions that are in the equation set. You can apply CSP in this problem but the time complexity will be $O({2}^n) $ where $n$ is the number of all variables in the same group. And also you can use sampling to get the number of mines which can satisfy all conditions with a success ratio.
Next, you can use the information on the number of total mines in the game to judge some groups' mine's numbers are impossible. The time complexity will be $O({2}^n) $ where $n$ is the number of all groups.  For many situations, the number of groups will be smaller than 5.

\subsection{Sample}
The last section has reduced the size of whole sample space but it still $O({2}^n) $, so we assume different groups are independent. We can sample for each group to get probabilities. For many situations, the number of variables in the same will be smaller than 25. We set the max number of samples are ${2}^18$ that we can run our program in an acceptable time about $5s$. To improve the utilization of samples we use the number of combinations to get probabilities of all samples' mines' number that we can get more acceptable samples.

\section{Result}
\subsection{Code and Presentation}
You can view our code in the github :
\href{https://github.com/TachikakaMin/MicroSoftBoom}{MineSweeper}
https://github.com/TachikakaMin/MicroSoftBoom

\subsection{Time}
\begin{figure}[htbp]
\centering
\includegraphics[width=6.5cm]{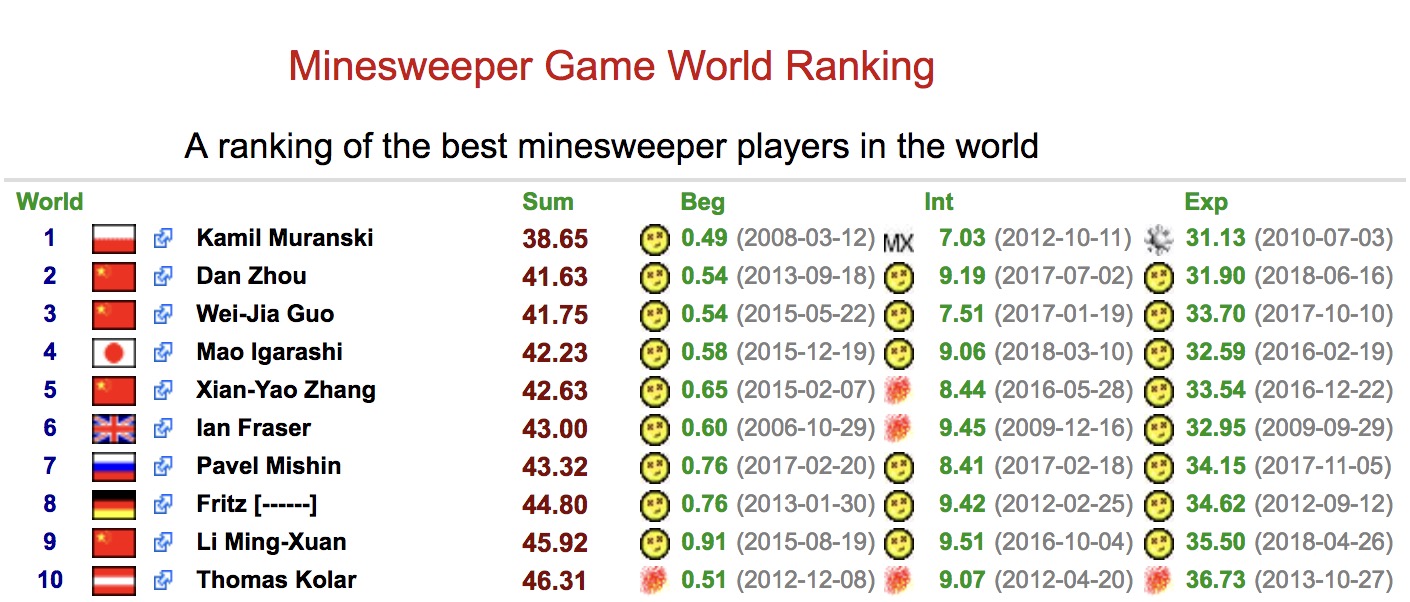}
\caption{Human Rank}
\label{rank}
\end{figure}

\begin{figure}[htbp]
\centering
\includegraphics[width=6.5cm]{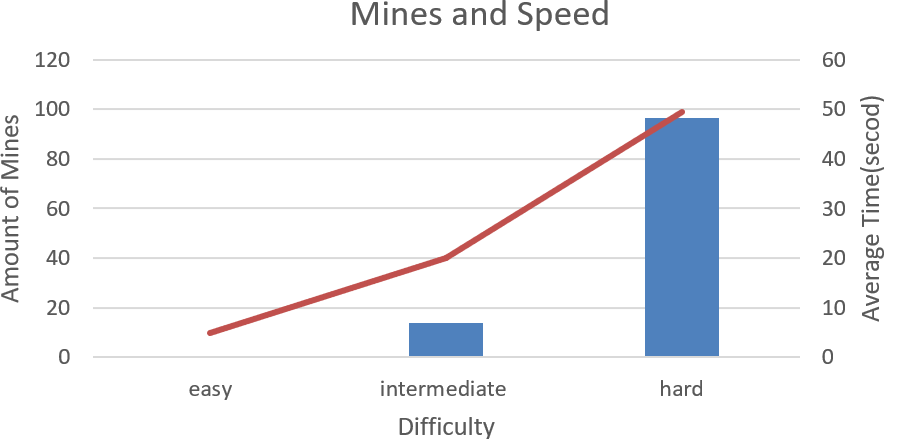}
\caption{Time spent by program}
\label{speed}
\end{figure}

This is a comparison of our program's running time and human time spent. Fig\ref{rank} shows China's Division of minesweeper, and Fig\ref{speed} shows the running time of our program under different difficult levels.

It is clear that our program is faster than human in simple and intermediate level. But there is no advantage in hard level. This is because humans only record the best score, while our score is average time taken by computers. And we know that minesweeper is a game which score has a large correlation with probability. Some situations can be easily solved, while others require our program to do a lot of calculations.

\subsection{Winning Rate}
\begin{figure}[htbp]
\centering
\includegraphics[width=8cm]{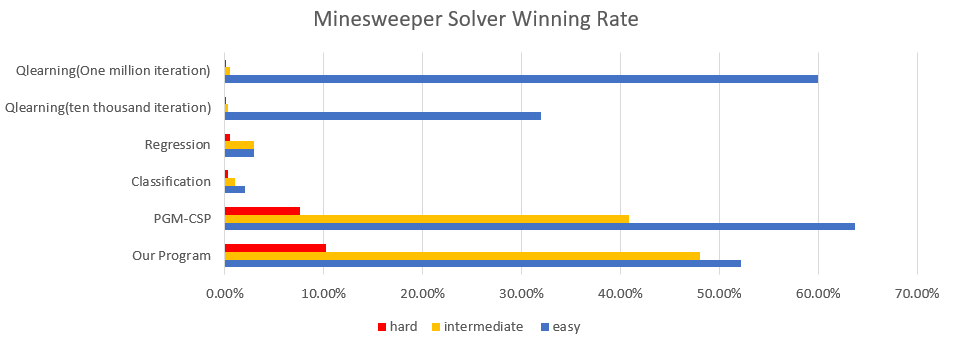}
\caption{Comparison with other algorithms}
\label{win}
\end{figure}

We used Google Scholar to search for other mine-sweeping algorithms
Fig\ref{win} shows a comparison of the winning rates of our algorithm and other paper algorithms.

From the above figure, we can see that our program has a great advantage over other algorithms in the difficult level and the inter. Especially the algorithms of some papers are almost impossible to win in the intermediate level.

So, why is our program slower in simple level? This is because, for other algorithms, the result of the first selecting block must not be a mine, while our program may be eliminated in the first step. On the simple level, the first step has a 15\% probability of stepping on a mine. Therefore, our program is not as good as other algorithms in simple level.

\section*{Acknowledgments}

The preparation of these instructions and the \LaTeX{} and Bib\TeX{}

\appendix

\bibliographystyle{named}
\bibliography{ijcai18}

\end{document}